\documentclass[letterpaper, 10 pt, conference]{ieeeconf}  
\usepackage[dvipsnames]{xcolor} 
\usepackage{graphicx} 
\usepackage{float}
\usepackage{booktabs}
\usepackage{multirow}
\usepackage{multicol}
\usepackage{wrapfig}
\usepackage{todonotes}
\usepackage{bm}
\usepackage{amsmath}
\let\labelindent\relax
\usepackage{enumitem}
\usepackage{xcolor}
\usepackage[normalem]{ulem}
\usepackage{hyperref}
\setlist{align=parleft,left=0pt..1em}
\usepackage{caption}

\renewcommand\vec{\mathbf}

\IEEEoverridecommandlockouts                              

\renewcommand\vec{\mathbf}

\usepackage[table]{xcolor}
\definecolor{lightred}{rgb}{0.988, 0.294, 0.0823}
\definecolor{lightblue}{rgb}{0.11, 0.541, 0.752}
\definecolor{lightgreen}{rgb}{0.3, 0.8, 0.3}
\usepackage[usenames,dvipsnames,table]{xcolor}

\hypersetup{
    colorlinks=true,
    linkcolor=MidnightBlue,
    citecolor=MidnightBlue,
    urlcolor=MidnightBlue,
    filecolor=MidnightBlue,
    bookmarks=true
}

\title{Visual-auditory Extrinsic Contact Estimation}

\author{Xili Yi$^*$$^{1}$, Jayjun Lee$^*$$^{1}$, Nima Fazeli$^{1}$
\thanks{* Equal contribution}
\thanks{$^{1}$ Robotics Department, University of Michigan, USA
        {\tt\small <yixili, jayjun, nfz>@umich.edu}}%
}

\newcommand{\papername}{\texttt{\textbf{VA2Contact}}}

\begin{document}
\maketitle

\begin{abstract}
Robust manipulation often hinges on a robot's ability to perceive extrinsic contacts—contacts between a grasped object and its surrounding environment. However, these contacts are difficult to observe through vision alone due to occlusions, limited resolution, and ambiguous near-contact states. In this paper, we propose a visual-auditory method for extrinsic contact estimation that integrates global scene information from vision with local contact cues obtained through active audio sensing. Our approach equips a robotic gripper with contact microphones and conduction speakers, enabling the system to emit and receive acoustic signals through the grasped object to detect external contacts. We train our perception pipeline entirely in simulation and zero-shot transfer to the real-world. To bridge the sim-to-real gap, we introduce a real-to-sim audio hallucination technique, injecting real-world audio samples into simulated scenes with ground-truth contact labels. The resulting multimodal model accurately estimates both the location and size of extrinsic contacts across a range of cluttered and occluded scenarios. We further demonstrate that explicit contact prediction significantly improves policy learning for downstream contact-rich manipulation tasks.

\end{abstract}

\section{Introduction}

\begin{figure}[h!]
\centering
\includegraphics[width=1.0\linewidth]{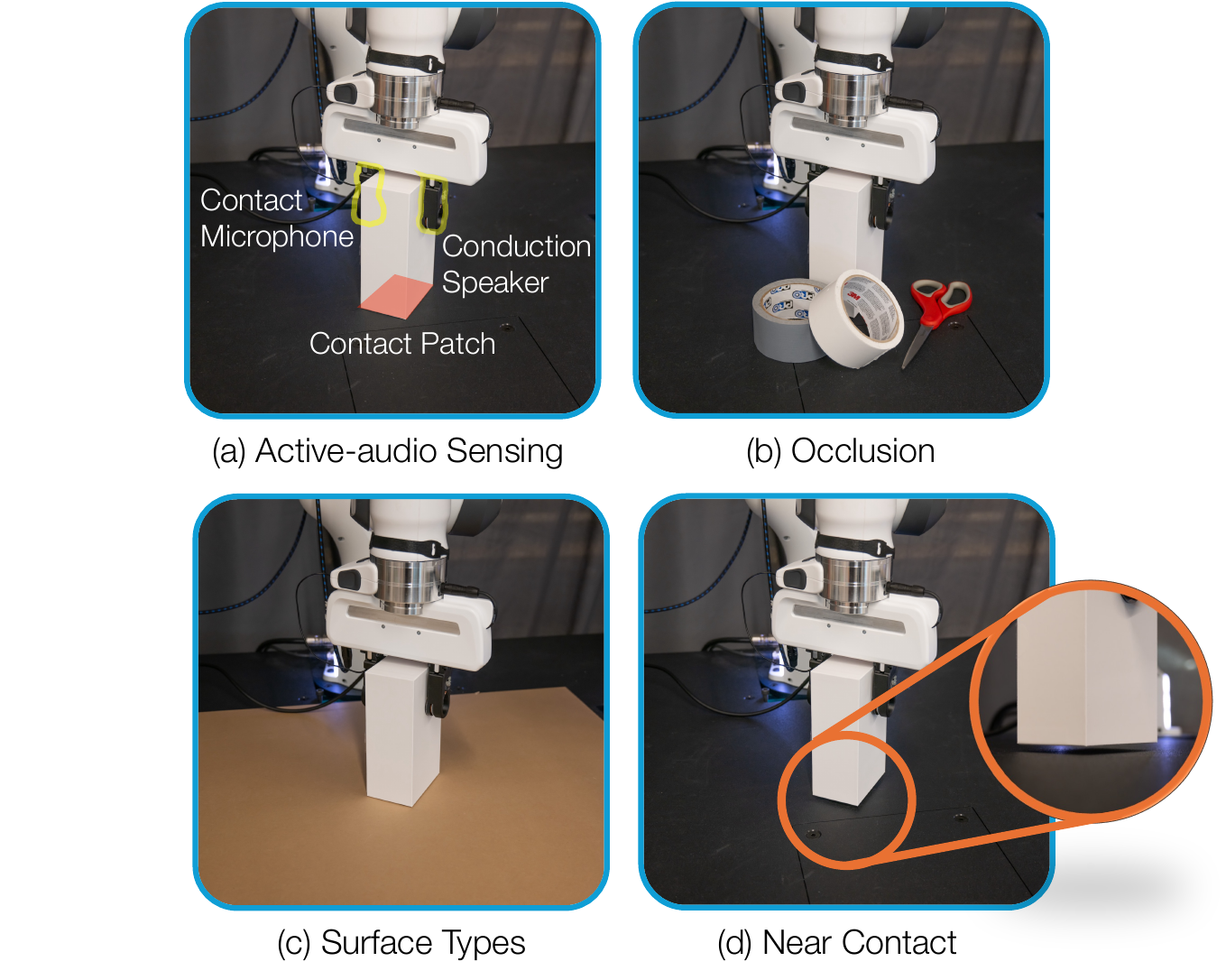}
\caption{(a) Our proposed fingers with an active conduction speaker and contact microphone emitting and receiving sound through the object. The absorption and reflection of the audio from the contact between the object and environment enables extrinsic contact estimation despite visual ambiguities. Challenges include (b) where objects occlude the contact between the box and the table, (c) where a different surface type can change the acoustic feedback, and (d) estimating the object's contact status for near-contact scenarios.}
\label{fig:teaser}
\end{figure}

Extrinsic contact estimation is a crucial capability for robots to accurately understand how tools interact with their environment. Properly perceiving these contacts enables the robot to plan and control its actions effectively. Vision-based methods allow observation of the entire scene, but they often fall short in providing sufficient local information, particularly when contacts are occluded or within the resolution limits of the sensor. While tactile sensors or force/torque sensors offer precise measurements of direct contact surfaces, they struggle to perceive indirect contacts, such as those between a tool and the environment. This challenge creates a sensing gap in extrinsic contact estimation. As illustrated in Fig.~\ref{fig:teaser}(d), distinguishing whether the objects are in contact when they are close to each other within the sensor’s resolution is difficult. Similarly, in Fig.~\ref{fig:teaser}(b), occlusions make it challenging to determine contact status as it is not directly observable.

To address this challenge, we propose a novel visual-auditory extrinsic contact estimation method, \papername{}. Our approach integrates global visual feedback with local information obtained through active audio sensing, as illustrated in Fig.~\ref{fig:teaser}(a). This enables the estimation of extrinsic contacts as masks in 2D image space, from which both the contact location and type can be inferred. Additionally, we introduce an innovative audio-hallucination technique to overcome the difficulty of scalably obtaining audio feedback in simulation. This technique involves injecting real-world audio data into the simulation dataset with corresponding labels, thereby bridging the sim2real gap.

Our work builds upon and improves recent advancements in the field. A closely related study is Im2Contact~\cite{im2contact}, where a single RGB-D camera was used to learn a probability map of contact from depth images, optical flow extracted from RGB images, and proprioception. While they demonstrated a promising ability to estimate contact locations in pixel space, two main issues persist. The first is that the existing method struggles with obtaining the location and shape of the contact patch in static scenes as it relies solely on the optical flow for extracting the temporal information of movement in the scene to infer contacts and such indication of motion is not always correlated to contact events. Second, the existing approach is limited by occlusions and the resolution of the camera. By incorporating local active-audio information, our approach provides more comprehensive contact details, regardless of the contact type, and in spite of heavy occlusion, addressing these limitations and enhancing the performance of binary extrinsic contact detection and the accuracy and robustness of estimating the geometry of the extrinsic contact patches.
\section{Related Works}
\label{sec:citations}
\noindent\textbf{Extrinsic Dexterity and Contact Estimation:}
In the field of manipulation, external contact sensing plays a crucial role. ``Intrinsic contact," which refers to the direct contact between the robot and the environment, has been well studied~\cite{de2006collision, manuelli2016localizing, bicchi1993contact}. 
However, when manipulating tools, estimating ``extrinsic contact," such as sensing the contact between a tool held by the robot and the environment, becomes important. 
Sensing extrinsic contact is more challenging due to the indirect transmission of contact force/torque and the uncertainties in the object's geometry, stiffness, and pose. 
\cite{higuera2023neural} combines neural fields and vision-based tactile sensing to estimate the probability of extrinsic contacts between object and environment for any point on the object's surface. 
Most studies rely on force/torque sensors or tactile sensors, which often involve strong assumptions or are limited to predefined contact configurations~\cite{yu2018realtime, ma2021extrinsic, kim2023simultaneous}. 
Some prior works require knowledge about the tool and the environment. 
For instance, \cite{sipos2022simultaneous} and the subsequent work~\cite{sipos2023multiscope} focused on estimating object pose and contact simultaneously from force/torque feedback. 
Additionally, data-driven methods have shown potential in estimating extrinsic contacts. \cite{im2contact} demonstrated that with visual data and robot proprioception, their model could predict the contact location in image space without assuming prior knowledge about the object and environment. 
\cite{kim2022active} uses active tactile feedback to regulate a more consistent contact mode and make contact estimates for peg insertion tasks. 
Recently, multimodal policies that combine proprioception, vision, and audio have also been explored to decompose tasks into stages~\cite{feng2024play} and by finetuning vision-language-action models~\cite{jones24fuse} for contact-rich manipulation tasks.

\noindent\textbf{Audio for Robotic Manipulation:}
Audio is also a widely used modality in the field of robotics. Due to its physical principles, audio signals can provide frequency-related characteristics. 
Passive acoustic sensing directly uses sound waves from structural vibrations~\cite{harrison2008scratch}. 
Some studies have modeled the sounds of object-surface interactions~\cite{lu2020towards}. 
This approach is also commonly used in soft pneumatic actuators (SPAs) to sense state changes. 
Previous studies also showed that in end-to-end robot learning algorithms, including audio signals as input improved task performance~\cite{thankaraj2023sounds, li2022see, mejia2024hearing, maniwav}. 
SonicSense~\cite{liu2024sonicsense} recently introduced a method to extract object material and shape information from in-hand acoustic vibrations, highlighting how such passive audio cues can enhance object understanding and contact estimation during manipulation.
Other works also use active audio sensing methods to gain more information from the environment. 
Zöller et al. detected contact by embedding a microphone into an SPA to measure the sound induced by contact \cite{zoller2018acoustic}. 
Similarly, studies~\cite{takaki2019acoustic, zoller2020active, mikogai2020contact, yoo2024poe} proposed embedding a microphone and speaker in an SPA, playing sweeping sounds, and measuring changes from the microphone to sense deformation or contacts. 
Multimodal sensing with audio can be used for surface proximity detection with piezoelectric transducers for robot collision avoidance~\cite{fan2021aurasenserobotcollisionavoidance}. 
There are also studies using active audio on rigid manipulator grippers to estimate grasping positions and for object recognition~\cite{lu2023active}. 
\cite{du2022play} uses a microphone attached at the gripper to receive audio feedback on contact events under partial observability. 

In this work, we leverage acoustic signals that transmit through solid objects by combining this with vision-based methods to estimate extrinsic contacts using a model-free approach. Unlike prior approaches that rely on fixed sensor setups or require assumptions about contact geometry, our method generalizes across contact types and tool configurations by fusing sound and vision in a flexible learning pipeline. Moreover, we introduce active-audio sensing to probe extrinsic contacts.


\section{Methodology}

\begin{figure*}[t!]
\centering
\includegraphics[width=\textwidth]{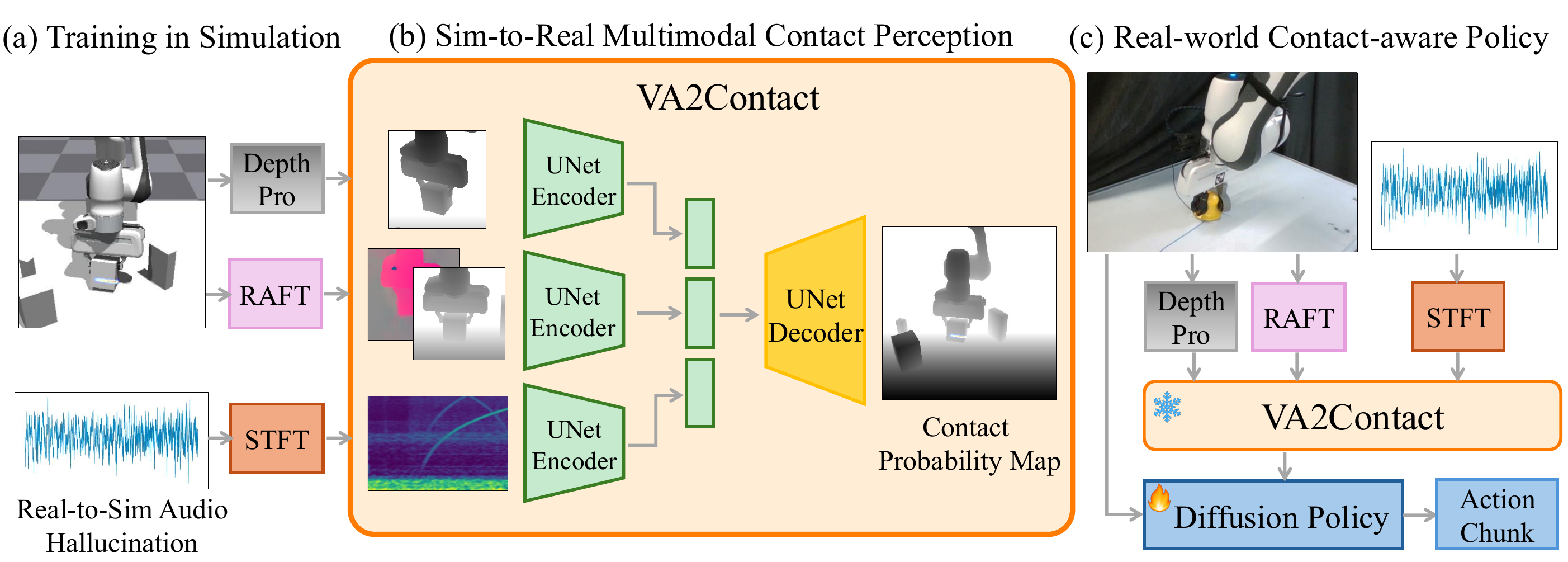}
\caption{\textbf{System Architecture.} \underline{V}isual-\underline{A}uditory Extrinsic \underline{Contact} Estimation (\papername{}) is trained in simulation with real-world audio collected through our active-audio sensing mechanism. Raw audio waveforms are processed with Short-Time Fourier Transform (STFT). \papername{} can zero-shot transfer to the real-world for contact prediction tasks. Here, the output contact probability map is overlaid onto the full depth image of the scene. Note the usage of off-the-shelf metric depth estimation models (Depth-Pro~\cite{bochkovskii2024depth}) and optical flow estimation model (RAFT~\cite{teed2020raft}) both for scalable sim-based training and real-world inference, to bridge the sim-to-real gap effectively. \papername{} unlocks contact perception under occlusions for contact-rich tool manipulation tasks such as wiping, which we demonstrate through real-world policy learning experiments.}  
\label{fig:overview}
\end{figure*}

\noindent\textbf{Problem Statement:} Consider a robot holding an unknown object and making contact with an unstructured environment. The goal of our method is to estimate the the location and shape of the \textit{extrinsic contact} between the grasped object and the environment. We assume no prior knowledge about the grasped object or environment. The inputs to our method are a depth map from a statically mounted camera, an optical flow image, fixed-length audio signatures measured at gripper finger tip, and proprioceptive sensing of the robot state. 

\noindent\textbf{Method Overview:} We train a multimodal model that can predict a per-pixel contact probability map over the scene given the inputs. All the training data except for audio are synthetically generated in simulation where pre-recorded audio data from the real-world are injected. The model is then zero-shot transferred to the real-world as in Figure~\ref{fig:overview}. In the following, we outline the active-audio sensing mechanism that enables extrinsic contact sensing beyond vision, data generation process to train such a model, the model architecture, and training details unique to this problem.

\begin{figure}[b]
\centering
\includegraphics[width=\linewidth]{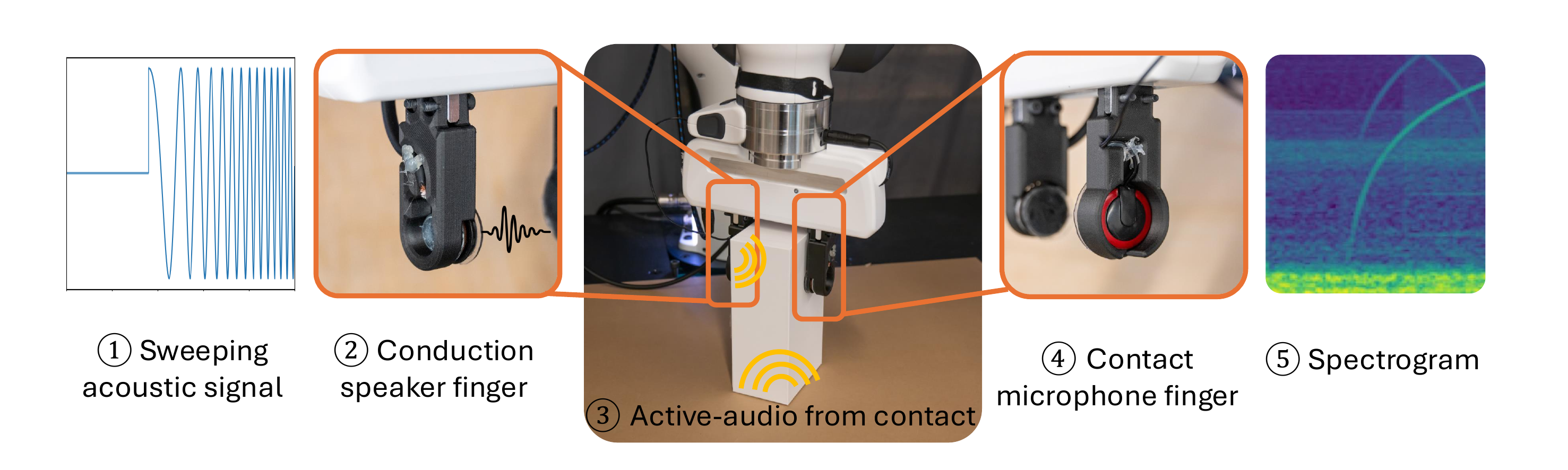}
  \caption{\textbf{Active-audio Sensing.} (1). A sweeping acoustic signal is generated from (2) using conduction speaker finger, where (3) the sound propagates through the object and vibrates with any extrinsic contact it makes, and (4) which is received at the contact microphone finger, and (5) the audio waveform is converted to a spectrogram.}
  \label{fig:audio}
\end{figure}

\noindent\textbf{Active-Audio Sensing:}
Suppose the robot is dragging the object it is grasping across a surface. The physical interaction between the object and environment creates an audio signature that contains a wealth of information about the interaction including its extrinsic contact patch size/shape, object/surface material types, contact location, and movement speed. However, if the object is in a static contact with the environment, there will be no audio feedback from the relative motion of the object, as audio signals require a source, either motion or an active audio source. To overcome this limitation and generalize to such static scenes, we introduce an active-audio system where one finger acts as an actuator (i.e. a speaker) by emitting controlled acoustic signals (i.e. a sweeping impulse signal), while the other finger as a receptor (i.e. a microphone) that receives the acoustic feedback, which changes depending on the grasped object's material properties and the presence of extrinsic contacts. As illustrated in Figure~\ref{fig:audio}, contact with the environment alters the received signal due to energy absorption, enabling active contact perception even in the absence of motion. This mechanism extends the robot’s perception to static contact scenarios, complementing vision and proprioception. Even without motion, contact with the environment alters the acoustic response by absorbing energy and modifying the transmission characteristics. This allows the system to distinguish between contact and non-contact scenarios based on the presence and properties of the received audio signal.

\noindent\textbf{Challenges in Data Generation:}
However, one major challenge in training a multimodal model for the task of visual-auditory extrinsic contact estimation is obtaining a dataset with aligned vision, audio, and ground-truth contact patch labels. In the real-world, it is relatively easy to obtain the audio information along side with both depth maps and robot proprioception data. However, obtaining correct contact patch masks is not only hard but also inaccurate with human labels, because there is no direct way to observe the exact contact patch shape and size. Meanwhile in simulation, it is simple to generate a large dataset with both correct contact patch masks and correct labels. However, obtaining realistic audio signals from simulation remains a great challenge. 

\noindent\textbf{Real-to-Sim Audio Hallucination:}
To overcome this challenge, we propose an audio hallucination technique. In the real-world, we collect audio signals ($\vec v_t$) as the robot manipulating various grasped objects by making diverse contact with the environment. During the robot motion, we record contact and contact geometry labels ($\vec g_t$) from the set \verb|[free, point, line, patch]|. We emphasize that we do not record contact locations, only the simple-to-observe contact labels. Thus the labeled dataset contains audio samples paired with contact types, represented as $\mathcal{D}_{\text{audio}}=\{(\vec v_0, \vec g_0), \cdots, (\vec v_n, \vec g_n)\}$. Next, in simulation we create scenes with the same robot and table as well as a large distribution of objects that the robot can grasp and interact with. This distribution of objects is not the same as the test objects used in the real-world setup. From the simulated visual scene, we obtain a depth map ($\vec d_t$) and obtain proprioceptive state ($\vec p_t$). We also take one depth map when the object is lifted without contact as a reference frame ($\vec r_t$) of the grasped object. Although the simulator can generate optical flow images, we generate the optical flow ($\vec f_t$) using an off-the-shelf RAFT model~\cite{teed2020raft} to later facilitate sim-to-real transfer, where we use the same model to generate the optical flow images for the real world test data and similarly use Depth-Pro~\cite{bochkovskii2024depth} for depth. It is noted that we generate optical flows only from depth maps to avoid the influence from shadows that would break the color constancy assumption. Most importantly, simulation also directly provides us with the contact geometry label and its corresponding shape ($\vec s_t$) when contact occurs. At each contact event, we randomly select a sample of the real-world audio corresponding to the label provided by the simulator and create a labeled dataset of contacts $\mathcal{D}=\{(\vec O_0, \vec s_0), \cdots, (\vec O_m, \vec s_m)\}$ where the first element is the observation vector $\vec O_i = (\vec v_i, \vec d_i, \vec r_i, \vec f_i, \vec p_i)$ and the second is the label.

The premise of our approach is that, while audio signatures vary across specific object interactions, they share consistent patterns that reflect the underlying contact geometry. A model trained on such data can learn to generalize by ignoring object-specific details and focusing on features indicative of contact type. For instance, point or line contacts tend to produce higher-pressure interactions, resulting in sharper, high-frequency sounds. In contrast, patch contacts generate a more diffuse sounds resembling white noise, with energy spread across a wider frequency range due to a larger contact area. 



\noindent\textbf{Model Architecture:}
For visual-auditory extrinsic contact estimation, our model architecture builds on the UNet architecture~\cite{unet} as illustrated in Figure~\ref{fig:overview}(b). The model consists of three streams of UNet encoders, each processing different streams of data. The first stream encodes a cropped depth image of the object at a reference frame. The second stream encodes a depth image and an optical flow image at the current frame, which are cropped around the projected end-effector pixel coordinates and stacked along the channel dimension. The third stream encodes an image of a log-mel spectrogram from a 1s long audio waveform sampled at 44.1 kHz. All input images are of shape $256 \times 256$ and the 7-DoF robot pose is fused at the bottleneck of the UNet. 

Additionally, a prior work \cite{im2contact} suggests a few strategies in which we adopt to facilitate better generalization and sim2real transfer. First, cropping the input images of the task scene centered around the end-effector position projected onto the pixel space simplifies the learning problem by removing background distractions and with an improved focus on the tool-environment interaction. For every cropping operation, additional three channels are stacked for coordinate convolution \cite{liu2018intriguingfailingconvolutionalneural} to provide the model with an explicit information about the spatial location of each pixel within the cropped image and help the model maintain the understanding of spatial relationships even after cropping. Another major challenge with the extrinsic contact estimation is when the grasped tool is heavily occluded in a cluttered scene that the parts of the grasped object that make contact with the environment is not visually observable. Thus, we opt to provide the model with a contact-free reference depth image of the object in the gripper. Moreover, visual ambiguities are present when differentiating an in-contact versus a near-contact state of the robot. This can be partially resolved by introducing the optical flow as an input to represent the temporal information of the robot motion. 

We further utilize the audio modality to acquire active feedback from extrinsic contact and interaction with the environment from the sine sweep impulse response that changes per contact mode type, grasped object, and robot motion. The output of the model is a contact probability map that can be thresholded to obtain the contact patch as a single channel mask image. 
\section{Implementation}

\noindent\textbf{Active-Audio Processing:}
The frequency characteristics of each object and contact vary depending on their physical properties, including stiffness and geometry. To capture these variations comprehensively, we transmit a sweeping impulse sound that spans a broad frequency range (20 to 20k Hz) through the object repeatedly at 1 Hz. We collect the audio feedback from a microphone and then use Short-Time Fourier Transform (STFT) and scaling to generate a log-mel spectrogram using 64 mel frequency bins. In the spectrogram, shown in Figure~\ref{fig:audio}, the x-axis represents time, the y-axis represents frequency, and the color represents energy at that time and frequency. The frequency axis is non-linearly scaled to show more information in the lower frequency range, which typically contains more useful information for contact. Given its ability to display both temporal and frequency information and their relationship, we choose this as the audio representation. We fix the duration of each audio sample to one second, ensuring that it captures the full impulse response across the entire sweeping frequency range. 

\noindent\textbf{Real World Setup:} 
For collecting and generating such scalable dataset for extrinsic contact perception, we use the following setup. In both simulation and the real-world, we use a 7-DoF Franka Emika Panda robot arm in a tabletop environment. Active-audio fingertips, illustrated in Fig.~\ref{fig:audio} are mounted onto the gripper, and both input and output audio signals are routed to a computer through a Focusrite Scarlett 4i4 3rd Gen amplifier. We set a single RealSense RGB-D camera pointing to the table for collecting visual feedback. We selected several objects from YCB dataset~\cite{calli2015ycb} (a spoon, clamp, tube, apple, lemon, orange, pear, screw driver, scissors, pods) and a few other objects such as a RealSense box and a dustpan as test objects. Note that these are not seen during training.

\noindent\textbf{Real World Audio Collection:}
To collect the active audio feedback with different types of contact, after the robot grasps an object, we teleoperate the robot to random orientation, and make contact with the environment. During the execution of each motion, we maintain the same contact mode(free, point, line, or patch contact), we also collect the active audio feedback from the scene, with sample rate of 44100 Hz. To provide the active audio signal, we input sweeping sound from 20Hz to 20000Hz in 1 second, and repeat it at 1 Hz. We also label the audio given the contact mode. We collected 1000 audio samples with variety of objects, contact shape and contact mode. 

\noindent\textbf{Simulation Setup and Real-to-Sim Audio Injection:}
We use the GPU-based Isaac Gym~\cite{isaacgym} simulation to replicate the real-world scene generating diverse extrinsic contact data at scale. In each episode, the robot grasps random objects and rotate to a random orientation, then lower down to the tabletop, and starts sliding the object on the table. For each environment, we randomly select cubes with random size and objects from ycb dataset\cite{calli2015ycb} that are smaller than 10cm in diameter. We obtain the contact points of the object and the environment, then project them back to the camera frame to obtain the contact masks. We also obtained a contact label from the size and shape of the contact, which are \verb|[free, point, line, patch]|. With these labels, we randomly select real-world audio samples with the same labels and then pair with depth image. The training data contains 6880 episodes where each episode contains around 60 frames, which approximates to a total of 440k data samples. 

\begin{figure}[H]
\centering
\includegraphics[width=\linewidth]{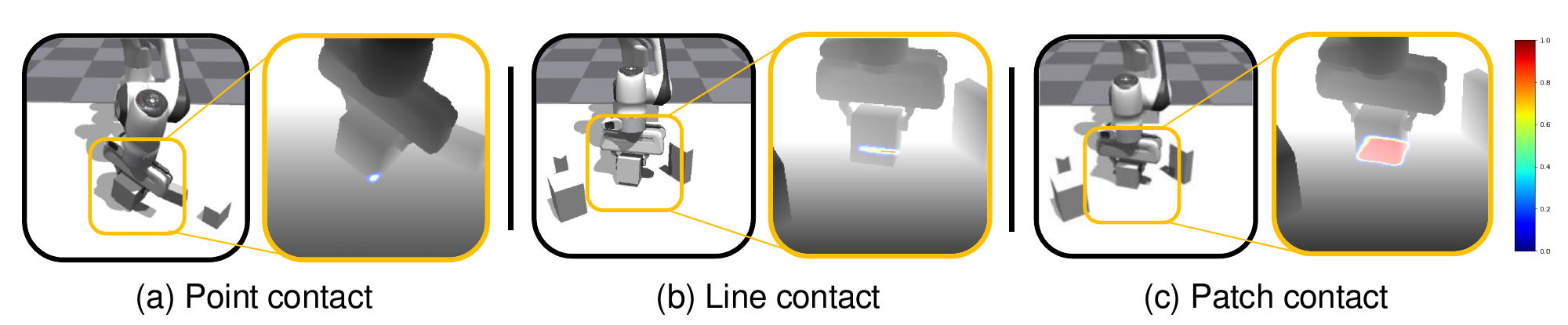}
\caption{\textbf{Different contact modes learned from simulation data.} The cropped depth is centered around the projected pixel coordinate of the EE pose. Sample predictions from test simulation data are shown as contact probability maps overlaid on top of scene depth.}
\label{fig:sim_results}
\end{figure}

\noindent\textbf{Training Details:}
We train our model with pixel-wise binary cross entropy loss with logits using the ground-truth contact masks obtained from simulation. We train for 20 epochs with a learning rate of $5e-4$ on a single A6000 GPU.  

\noindent\textbf{Real World Test Data Collection:}
To evaluate \papername{} on extrinsic contact detection, we collect data that consists of RGB-D images of the scene, EE pose, camera extrinsics, a reference RGB-D image, audio feedback, and a manually labelled extrinsic contact mask. We use the same off-the-shelf models (Depth-Pro and RAFT) to obtain consistent depth and optical flow images 

Additionally, for collecting the real-world depth images, we primarily use the monocular metric depth estimation model Depth-Pro~\cite{bochkovskii2024depth} to generate hole-less depth map from raw RGB images of the scene. We collected data in 4 cases: general, occlusion, near-contact, and different surface cases. We use RAFT~\cite{teed2020raft} to generate optical flow images. We use the same audio collection pipeline as audio collection for test data collection. 

For downstream policy learning with \papername{} in the perception pipeline, we use an Oculus Quest 2 device to teleoperate and collect 41 demonstrations for the wiping task that amount to 16k timesteps of training data with audio feedback at 15 Hz. We preprocess this data to generate consistent depth, optical flow, and contact maps using Depth-Pro, RAFT, and \papername{} to train diffusion policy that takes two images (one for raw RGB and one for contact prediction image) and joint state as input. The diffusion policy is trained on relative joint action space that predicts a action horizon of 16 where we execute the first 8 for receding horizon control at 15 Hz.


\section{Experiments}
\label{sec:experiments}

\begin{table*}[t]
\centering
\scalebox{0.9}{
\begin{tabular}{lccccccc}
\toprule
\textbf{Models} & \multicolumn{4}{c}{\textbf{Binary Contact Detection (macro)}} & \multicolumn{3}{c}{\textbf{Contact Location, Shape, \& Size (macro)}} \\
\cmidrule(lr){2-5} \cmidrule(lr){6-8}
 & \textbf{Prsn $\uparrow$} & \textbf{Rcll $\uparrow$} & \textbf{F1 $\uparrow$} & \textbf{Acc $\uparrow$} & \textbf{BCE $\downarrow$} & \textbf{IOU $\uparrow$} & \textbf{CD ($\mathrm{px}^2$) $\downarrow$} \\
\midrule
Im2Contact          & 0.87 & 0.87 & 0.86 & 0.58 & \textbf{0.031} & 0.092 & 15.01 \\
\papername{} w/o flow & \textbf{0.99} & 0.89 & 0.94 & 0.91 & 0.038 & 0.096 & 11.73 \\
\papername{}          & 0.94 & \textbf{0.95} & \textbf{0.94} & \textbf{0.93} & 0.038 & \textbf{0.240} & \textbf{6.44} \\
\bottomrule
\end{tabular}
}
\caption{\textbf{Real world results for sim-to-real extrinsic contact estimation on all 4 scenarios.} Models are evaluated on binary contact detection and the accuracy of contact geometry estimation, which is only provided for the true positive cases where the model predicts a mask above a threshold level and the ground-truth is in-contact. CD: Chamfer Distance. IOU: Intersection over Union. BCE: Binary cross-entropy.}
\label{tab:general}
\end{table*}

\begin{table*}[h!]
\centering
\scalebox{0.95}{
\begin{tabular}{llccccccc}
\toprule
\textbf{Subset} & \textbf{Model} &
\textbf{Prsn $\uparrow$} & \textbf{Rcll $\uparrow$} & \textbf{F1 $\uparrow$} &
\textbf{Acc $\uparrow$} & \textbf{BCE $\downarrow$} &
\textbf{IOU $\uparrow$} & \textbf{CD ($\mathrm{px}^2$) $\downarrow$} \\
\midrule
\multirow{3}{*}{\textbf{S2: Surface Types}}
& Im2Contact           & \textbf{1.00} & 0.78 & 0.88 & 0.78 & 0.040 & 0.139 & 10.52 \\
& \papername{} w/o flow & \textbf{1.00} & 0.86 & 0.92 & 0.86 & 0.060 & 0.116 & 10.74 \\
& \papername{}          & \textbf{1.00} & \textbf{0.94} & \textbf{0.97} & \textbf{0.94} & 0.054 & \textbf{0.221} & \textbf{4.36} \\
\midrule
\multicolumn{9}{l}{\textbf{S3: Near Contact (no positive GT; Precision/Recall/F1 not defined)}} \\
\midrule
& Im2Contact           & \multicolumn{3}{c}{TN=2,\; FP=48}  & 0.04 & -- & -- & -- \\
& \papername{} w/o flow& \multicolumn{3}{c}{TN=48,\; FP=2}  & 0.96 & -- & -- & -- \\
& \papername{}         & \multicolumn{3}{c}{TN=49,\; FP=1}  & \textbf{0.98} & -- & -- & -- \\
\midrule
\multirow{3}{*}{\textbf{S4: Occlusions}}
& Im2Contact           & 0.91 & 0.89 & 0.90 & 0.82 & 0.029 & 0.079 & 15.30 \\
& \papername{} w/o flow& \textbf{1.00} & 0.87 & 0.93 & 0.88 & \textbf{0.025} & 0.099 & 10.84 \\
& \papername{}         & 0.96 & \textbf{1.00} & \textbf{0.98} & \textbf{0.96} & 0.034 & \textbf{0.220} & \textbf{10.16} \\
\bottomrule
\end{tabular}
}
\caption{\textbf{Real-world results for sim-to-real extrinsic contact estimation.}  
Binary detection metrics (Precision/Recall/F1/Accuracy) are reported where positive ground-truth contacts exist (S2, S4).  
For S3, only near-contact negatives are present; we therefore report true negatives (TN), false positives (FP), and accuracy.  
Geometry metrics (BCE, IOU, Chamfer Distance) are evaluated on true positives only.}
\label{tab:combined_results}
\footnotetext{Precision, Recall, and F1 are undefined when no positive ground-truth contacts occur.}
\end{table*}

\subsection{Extrinsic Contact Estimation}
\noindent\textbf{Baselines and Ablations:} We compare our visual-auditory method against a similar model-free and vision-only method Im2Contact~\cite{im2contact}. We also evaluate a variant of \papername{} trained without optical flow to experiment if audio can fully replace it.
\noindent\textbf{Simulation Test Scenarios:}
We first evaluate our methods in simulation 

\noindent\textbf{Real World Test Scenarios:}
To evaluate the sim2real transfer of our models on the extrinsic contact estimation problem, we design the following real world test scenarios. These scenarios highlight the primary challenges, including demonstrating sim2real transfer under substantial sim2real gap (S1, S2), contacts invisible to vision (S3, S4).
\begin{figure*}[t]
\centering
\includegraphics[width=0.96\linewidth]{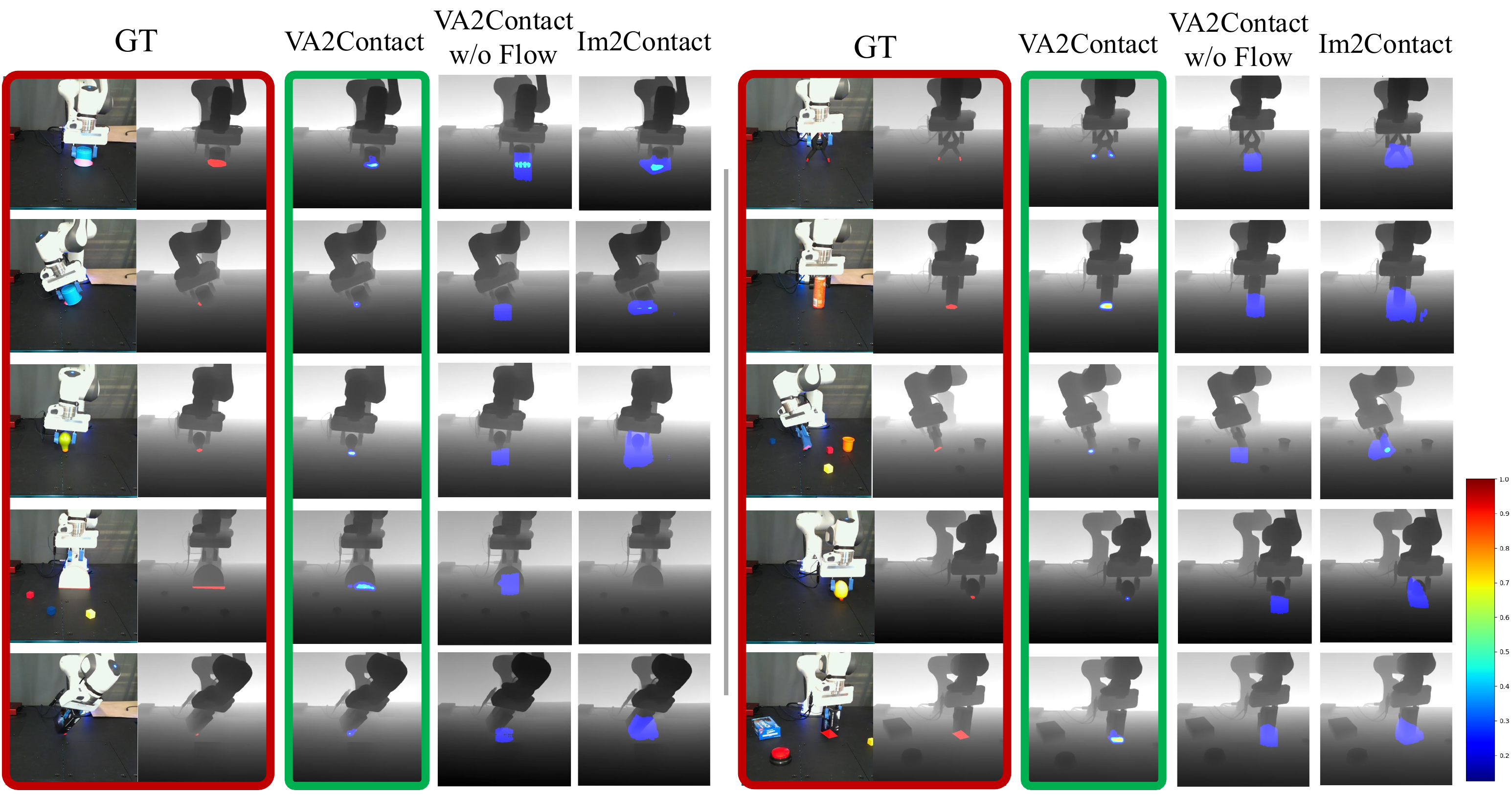}
\caption{\textbf{Sim-to-Real Transfer and Real-world Extrinsic Contact Predictions for S1.} The RGB-D images with GT contact probability masks are wrapped in \textcolor{red}{red}. Three models, \papername{}, \papername{} w/o optical flow, and Im2Contact, are tested where the results are shown per column. \papername{}'s contact probability predictions are overlaid to depth images, wrapped in \textcolor{lightgreen}{green}. The color bar represents contact probability (\textcolor{blue}{0.0 $\leftarrow$} \textbar{} \textcolor{red}{$\rightarrow$ 1.0}). All grasped objects used for real-world testing are unseen geometries (a cup, pear, dustpan, box, blue sponge, lemon, can, clamp). \papername{} is able to zero-shot predict diverse contact types over objects with varying properties.}
\label{fig:showcase1}
\end{figure*}

\begin{figure*}[ht!]
\centering
\includegraphics[width=\linewidth]{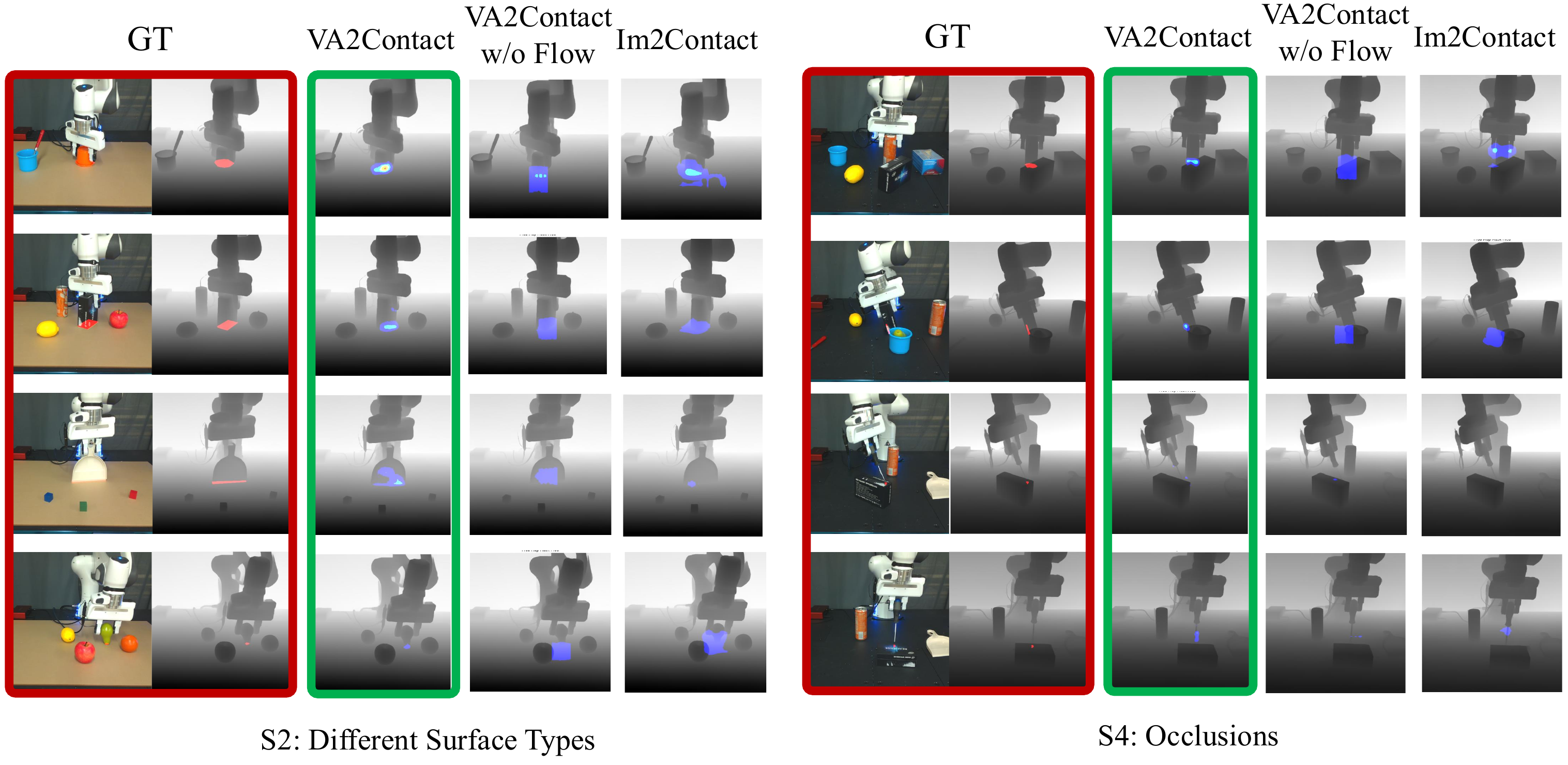}
\caption{\textbf{Sim-to-Real Transfer and Real-world Extrinsic Contact Predictions for S2 and S4.} The RGB-D images with GT contact probability masks are wrapped in \textcolor{red}{red}. Three models, \papername{}, \papername{} w/o optical flow, and Im2Contact, are tested where the results are shown per column. \papername{}'s contact probability predictions are overlaid to depth images, wrapped in \textcolor{lightgreen}{green}. The color bar represents contact probability (\textcolor{blue}{0.0 $\leftarrow$} \textbar{} \textcolor{red}{$\rightarrow$ 1.0}). All grasped objects used for real-world testing are unseen geometries (a cup, pear, dustpan, box, blue sponge, lemon, can, clamp). \papername{} is able to zero-shot predict diverse contact types over objects with varying properties.}
\label{fig:showcase2}
\end{figure*}

\begin{itemize}
\itemsep0em 
    \item \textbf{S1}: Generalization to unseen gripper-held object types and geometry. (Figure~\ref{fig:showcase1})
    \item \textbf{S2}: Generalization to unseen table surface to introduce audio variation. (Figure~\ref{fig:showcase2})
    \item \textbf{S3}: Near contact cases that are not in contact. 
    \item \textbf{S4}: Cases where the contact between gripper-held object and the table is occluded. (Figure~\ref{fig:showcase2})
\end{itemize}
In total, we collected 310 evaluation samples. 
The number per contact mode is \texttt{free:point:line:patch} = \{48:37:41:34\} for \textbf{S1}, 
\{0:15:15:20\} for \textbf{S2}, 
\{4:14:9:23\} for \textbf{S3}, 
and \{50:0:0:0\} for \textbf{S4}.

\noindent\textbf{Experimental Results:}
\label{sec:results}
The main results of the sim2real transfer of visual-auditory extrinsic contact estimation in the real world setup are shown in Table~\ref{tab:general} and Figure~\ref{fig:showcase1}. Here we show the performance of our methods across all 4 scenarios.  In the \textit{general} case, the model with audio overall exhibits higher recall and F1 score, indicating a superior ability to detect true contacts. This suggests that audio cues help reduce false negatives without substantially increasing false positives. 
\papername{} w/o optical flow scores the highest across all metrics for binary contact detection because audio provides a more direct information about contact whereas optical flow just tells us where things moved, which might not be directly correlated. However, in terms of the actual contact patch prediction, \papername{} which uses both active-audio and optical flow performs the best as shown in Figure~\ref{fig:showcase1}.

We also explore edge cases such as visual occlusions and near-contact scenarios where active-acoustic signals provide critical cues,  as shown in Figure~\ref{fig:showcase2}. 
We evaluate \papername{} under three challenging real-world settings: \textit{different surface types}, \textit{near-contact}, and \textit{occlusions}. On different surfaces (S2, Table~\ref{tab:combined_results}), both vision-only and audio-augmented models achieve strong binary contact detection. While IOU is slightly better without audio, our method generalizes well despite variation in acoustic feedback, showing audio input remains effective under domain shifts. In near-contact scenarios (S3, Table~\ref{tab:combined_results}), \papername{} significantly reduces false positives compared to Im2Contact, indicating its strength in disambiguating ambiguous visual cases via active sensing. Under occlusions (S4, Table~\ref{tab:combined_results}), \papername{} consistently outperforms all baselines in both detection and geometry metrics (e.g., Recall: 0.94 vs. 0.78; F1: 0.97 vs. 0.88), demonstrating that audio cues robustly compensate for missing visual information.

\subsection{Contact-aware Policy Learning}

\begin{figure}[]
    \centering
    \includegraphics[width=0.8\linewidth]{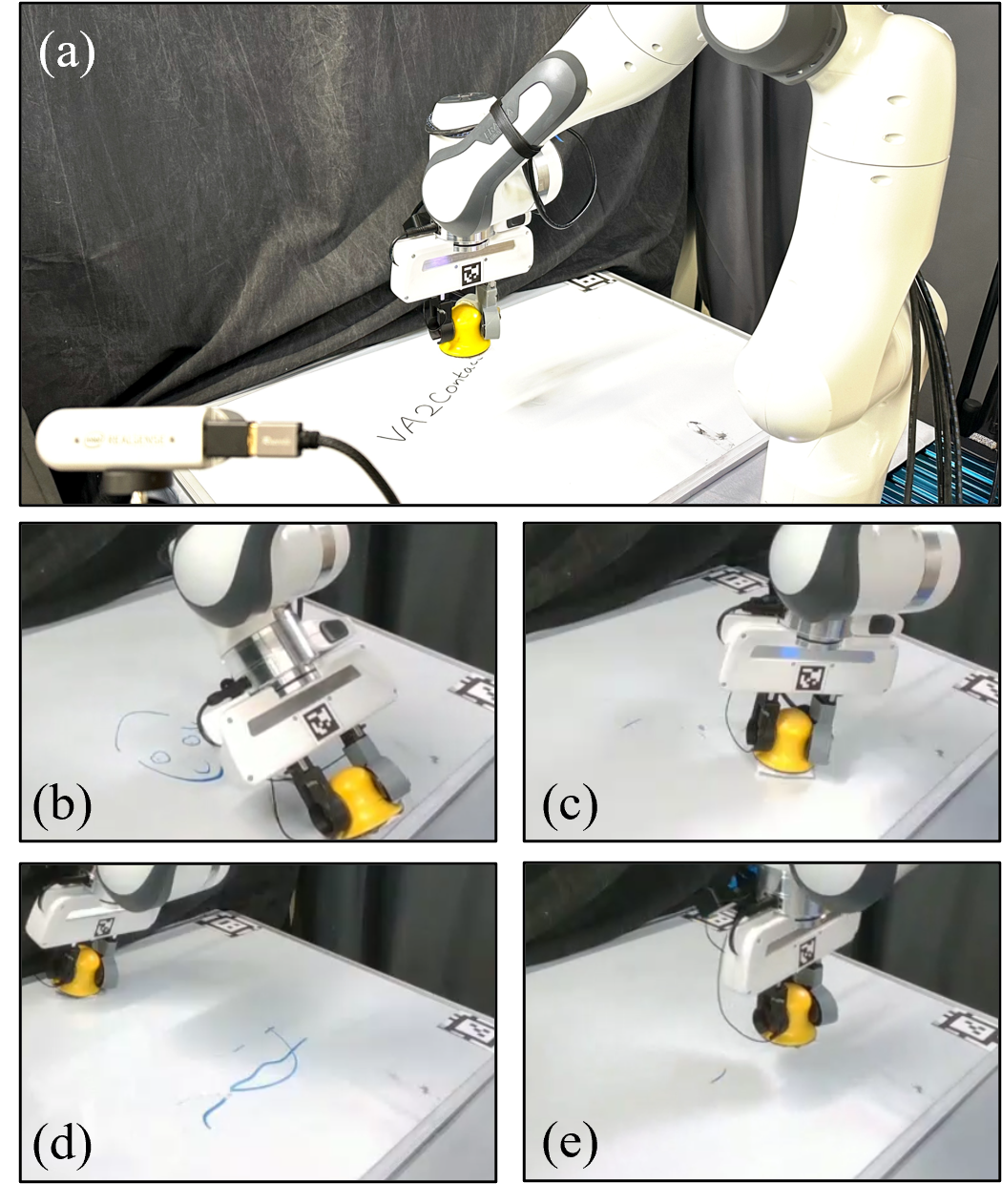}
    \caption{Real-world wiping task using a contact-aware diffusion policy enabled by \papername{}.
(a) Experimental setup where the robot wipes a whiteboard using our method.
(b, c) Comparison of baseline and \papername{} wiping a drawn face.
(d, e) Comparison of baseline and \papername{} wiping a boat.
In both tasks, \papername{} enables more consistent and complete contact with the surface, leading to better task performance.}
    \label{fig:wiping}
\end{figure}

Given this sim2real visual-auditory extrinsic contact sensing capability, we evaluate whether this can improve the downstream policy learning and performance for contact-rich and occlusion-heavy manipulation tasks. We selected a wiping task, which requires the robot to hold a marker eraser and maintain contact to erase the markers, as shown in Figure \ref{fig:wiping}. To test this, we train a baseline diffusion policy with just an RGB image stream and a contact-aware diffusion policy as in Figure~\ref{fig:overview} that takes the contact probability map image in addition to the RGB image stream as inputs. We evaluate the performances over 10 rollouts for each model. The baseline model achieved 4/10 success due to inconsistent contact with the whiteboard whereas the contact-aware policy achieved 8/10 success rate. 
Overall, the integration of audio input into the contact estimation model enhances its ability to reliably detect and localize contacts in complex real-world environments and bridges the sim-to-real transfer well. The audio cues effectively complement visual data, particularly in situations where visual information is incomplete or ambiguous such as under occlusions, which are common in real world manipulation tasks. Moreover, we find that such multimodal contact-aware representations can enhance contact-rich manipulation capabilites further.




\section{Conclusion and Discussions} 
\label{sec:conclusion}

This paper presented \papername{}, a simulation-trained method for estimating extrinsic contacts using multimodal inputs: fingertip audio, vision, and proprioception. Active audio sensing provides local cues that complement vision, enabling robust contact perception under occlusions and ambiguity. To bypass the challenge of simulating audio, we introduce a real-to-sim hallucination strategy that injects real audio into training, enabling strong zero-shot transfer to the real world and outperforming visual-only baselines.

Despite promising results, limitations remain. Internal robot vibrations, especially from the Franka gripper, reduce audio signal quality. Manual annotations of contact masks may be biased due to occlusion. Our current audio representation may not fully capture rich contact cues—future work may explore pretrained models. Finally, the dataset focuses on rigid objects; generalizing to deformable or liquid-filled items and contact chains will require better sensing and annotation strategies.

\bibliographystyle{unsrt}
\bibliography{example}  
\newpage

\end{document}